%% file: arxiv.tex
\documentclass[10pt,twocolumn,letterpaper]{article}

\usepackage[pagenumbers]{iccv} 

\input{preamble}

\definecolor{iccvblue}{rgb}{0.21,0.49,0.74}

\usepackage[pagebackref,breaklinks,colorlinks,allcolors=iccvblue]{hyperref}

\title{Few-shot Human Action Anomaly Detection via \\a Unified Contrastive Learning Framework}

\author{Koichiro Kamide$^{1}$, Shunsuke Sakai$^{2}$, Shun Maeda$^{2}$, Chunzhi Gu$^{2}$, Chao Zhang$^{1}$ \\
$^{1}$University of Toyama, 
$^{2}$University of Fukui\\
}

\begin{document}
\maketitle
\begin{abstract}
Human Action Anomaly Detection (HAAD) aims to identify anomalous actions given only normal action data during training.
Existing methods typically follow a one-model-per-category paradigm, requiring separate training for each action category and a large number of normal samples. 
These constraints hinder scalability and limit applicability in real-world scenarios, where data is often scarce or novel categories frequently appear.
To address these limitations, we propose a unified framework for HAAD that is compatible with few-shot scenarios. 
Our method constructs a category-agnostic representation space via contrastive learning, enabling AD by comparing test samples with a given small set of normal examples (referred to as the support set).
To improve inter-category generalization and intra-category robustness, we introduce a generative motion augmentation strategy harnessing a diffusion-based foundation model for creating diverse and realistic training samples. Notably, to the best of our knowledge, our work is the first to introduce such a strategy specifically tailored to enhance contrastive learning for action AD.
Extensive experiments on the HumanAct12 dataset demonstrate the state-of-the-art effectiveness of our approach under both seen and unseen category settings, regarding training efficiency and model scalability  
for few-shot HAAD.
\end{abstract}

\section{Introduction}\label{intro}
\begin{figure*}[tb]
    \centering
    \includegraphics[width=0.7\linewidth]{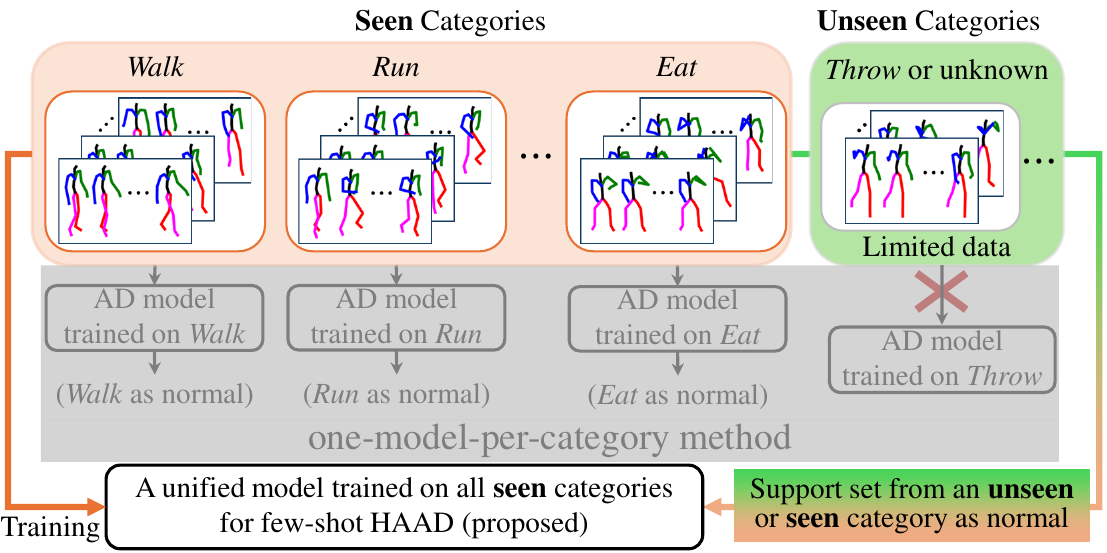}
    \caption{\textbf{Comparison between vanilla AD (one-model-per-category) and our unified few-shot AD approach on the task of HAAD.} Vanilla methods require training an individual model per category with large amounts of data, which is less practical when applying to unseen categories. Our method trains a single model on seen categories and allows for both category-dependent and category-agnostic few-shot AD, by using a small support set that flexibly defines the normal category.}
    \label{fig:problem-setting}
\end{figure*}
Anomaly Detection (AD) is a fundamental task in machine learning that aims to identify patterns deviating from the normality in terms of appearance, structure, or behavior. This task plays a critical role in various real-world applications such as industrial defect inspection~\cite{bergmann2020uninformed, Bergmann_2019_CVPR, defard2021padim, kahler2022anomaly, baitieva2024supervised}, medical diagnosis~\cite{schlegl2019f, han2021madgan, zhang2022unsupervised, kim2023unsupervised}, and autonomous driving~\cite{xiao2025every, basart2022scaling, nekrasov2025spotting}, where detecting unexpected events is essential for ensuring safety and reliability.
Among these diverse applications, this paper focuses on human actions. In particular, we adopt the setting of Human Action AD (HAAD), a task recently introduced by S. Maeda~\cite{maeda2024frequency}, in which one specific action category is designated as normal (\textit{e.g., Walk}) and all others (\textit{e.g., Run, Sit}) are treated as anomalies. HAAD is expected to play an important role in safety-critical applications such as industrial monitoring, crime prevention, and elderly care by enabling early detection of anomalous behaviors.

Due to the difficulty of collecting diverse anomalous actions, this task is generally tackled in the unsupervised setting, where solely normal data is presented in the training set. Also, as illustrated by the gray-shaded region in Fig.~\ref{fig:problem-setting}, existing HAAD methods~\cite{flaborea2023multimodal,hirschorn2023normalizing,maeda2024frequency} follow a one-model-per-category paradigm, where a separate model is trained for each action category. This paradigm demonstrates promising performance when sufficient normal training data is available. 
However, this approach faces two critical limitations: the requirement of retraining for each category and a strong dependency on large volumes of normal data. The need to retrain models introduces significant computational costs, increases model management overhead, and results in poor scalability in real-world deployment settings. Moreover, in practical scenarios, it should be noticed that the number of available samples per category is often highly limited. When only a small amount of data is available for certain action categories, training reliable models becomes infeasible. These limitations significantly hinder the practicality and scalability of existing methods, especially in environments where data is scarce or heterogeneous.

\begin{figure*}[tb]
    \centering
    \includegraphics[width=0.9\linewidth]{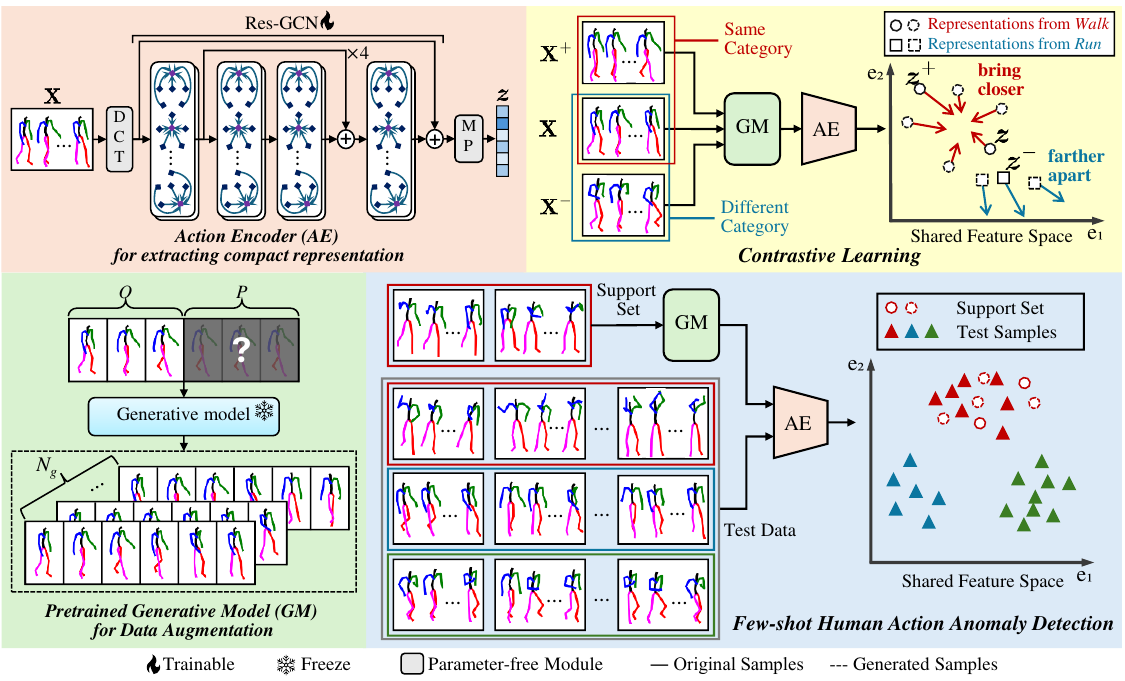}
    \caption{\textbf{Overview} of our unified modeling for few-shot HAAD.}
    \label{fig:overview}
\end{figure*}

To overcome the above limitations of retraining and the data-hungry issue, we propose a unified contrastive learning-based framework for HAAD that is compatible with few-shot scenarios. An overview of the proposed approach and framework is illustrated in Fig.~\ref{fig:problem-setting} and Fig.~\ref{fig:overview}, respectively. Inspired by the success of multi-category AD~\cite{you2022unified, huang2022registration, Zhao_2023_CVPR, lu2023hierarchicalvectorquantizedtransformer, yin2023lafitelatentdiffusionmodel, he2024mambaad, gao2024learning, guo2025dinomaly, beizaee2025correcting}, our core idea is to construct a category-agnostic representation space where actions from different categories are embedded into a single shared space with a discriminative characteristic. Importantly, this shared discriminative space eliminates the need to train separate models and enables HAAD by measuring the similarity between a test action and a few normal examples, often referred to as a support set.

To embed the actions into such a representation space, we design our model as a contrastive action encoder based on Graph Convolutional Networks (GCN). Specifically, the motion sequences are first encoded into the frequency space using the Discrete Cosine Transform (DCT) to retain only the meaningful low-frequency components, followed by a Residual GCN (Res-GCN) to yield the compact embedding.
During training, inspired by the previous few-shot framework ~\cite{huang2022registration}, we adopt contrastive learning~\cite{chen2020simple} using training data from multiple categories. This training design enables the encoder to construct a shared feature space in which actions with similar semantics become closer while dissimilar ones are repelled farther apart.
Eventually, the anomaly scoring is achieved by computing the average similarity between the encoded test sample and all reference samples in the support set.

Additionally, we introduce a generative motion augmentation strategy for improving inter-category discrimination and robustness to intra-category variation. In image-based contrastive learning~\cite{kim2023unsupervised, chen2020simple, chen2021exploring, huang2022registration, huynh2022boosting}, data augmentation is typically achieved through hand-crafted transformations such as cropping or flipping. However, such conventional transformations are not directly applicable to skeletal pose sequences, where both the spatial relationships among joints and temporal continuity across frames are crucial for maintaining the semantic consistency.
Therefore, we employ HumanMAC~\cite{Chen_2023_ICCV}, which is a diffusion-based generative model specifically designed for human motion. In particular, HumanMAC formulates motion prediction as a masked motion completion task in the frequency domain. Given an observed sequence, HumanMAC predicts multiple plausible future motions, enabling the generation of semantically consistent and structurally valid motion variants. Due to our generative augmentation, our model can be empowered to better characterize discriminative motion representations to boost detection accuracy.

In conclusion, our main contributions are summarized as follows:
\begin{itemize}[itemsep=0pt, parsep=0pt]
    \item We extend a unified framework for HAAD, which is compatible with few-shot scenarios and does not require category-specific retraining.
    \item We introduce a generative model to augment the training set and diversify support samples during inference. To the best of our knowledge, this is the first work to use generative modeling specifically to enhance contrastive learning for HAAD.
    \item We enable generalization to novel action categories without retraining, achieved through a category-agnostic representation space and similarity-based scoring defined by a few normal samples.
    \item Experimental results on the HumanAct12~\cite{guo2020action2motion} dataset validate the effectiveness of our approach under both seen and unseen category settings, achieving an average AUC of 86.9\%, which is 2.6\% higher than the previous state-of-the-art~\cite{maeda2024frequency}.
\end{itemize}

\section{Related work}

\subsection{Anomaly detection in human actions}
AD in human actions aims to identify unexpected behaviors from recorded videos~\cite{li2013anomaly, sabokrou2018deep, wang2018video, morais2019learning, ramachandra2020survey, georgescu2021anomaly, georgescu2021background}.
The main challenge in this problem is that prior models are often sensitive to background variations, lighting changes, or camera viewpoints, which can obscure the true semantics of human motion.
To tackle this challenge, skeleton-based representations~\cite{du2015hierarchical, yan2018spatial, shi2019skeleton, duan2022revisiting, ren2024survey} have gained attention for providing a compact and robust description of human actions by focusing solely on joint movements, making them well-suited for HAAD tasks.

Recent skeleton-based AD methods can be categorized into reconstruction-based~\cite{flaborea2023multimodal} and normalizing flow-based~\cite{hirschorn2023normalizing, maeda2024frequency} frameworks. 
Flaborea \textit{et al}.~\cite{flaborea2023multimodal} proposed a reconstruction-based HAAD framework leveraging diffusion models. Motion sequences are encoded into latent features and reconstructed via a denoising decoder, with reconstruction error used as the anomaly score. 
Hirschorn \textit{et al}.~\cite{hirschorn2023normalizing} developed a normalizing flow-based method that learns an invertible mapping from action space to a simple latent distribution (\textit{e.g.}, Gaussian), assigning high likelihood to normal actions and regarding actions with low likelihood as anomalies.
Maeda \textit{et al}.~\cite{maeda2024frequency} proposed a frequency-domain learning strategy that extracts upper and lower body features separately to capture both global and local motion patterns. Anomaly scores are computed during inference by applying k-nearest neighbors (k-NN) to the learned features.
While these methods achieve strong category-specific performance, they follow a one-model-per-category paradigm, requiring separate training and substantial normal samples for each category. 
Furthermore, both reconstruction-based and normalizing flow-based methods are not suited for the few-shot scenario. To address this, our work introduces a unified framework that constructs a category-agnostic representation space, enabling few-shot HAAD without retraining and maintaining high accuracy even with limited normal samples. We provide a comparison between previous HAAD methods and our proposed framework in Tab. ~\ref{tab:comp-framework}.
\begin{table}[tb]
\centering
\caption{\textbf{Comparison of different frameworks.} 
\textit{Recon}, \textit{NF}, and \textit{CL} denote reconstruction-, normalizing flow-, and contrastive learning-methods, respectively. \textit{Single-cat.} and \textit{Multi-cat.} refer to training with normal data from a single category and multiple categories, respectively.}
\small
    \begin{tabular}{lccc}
    \toprule
    Method & framework & Training & few-shot \\
    \midrule
    MoCoDAD~\cite{flaborea2023multimodal} & \textit{Recon} & \textit{Single-cat.} & \xmark \\
    STG-NF~\cite{hirschorn2023normalizing} & \textit{NF} & \textit{Single-cat.} & \xmark \\
    MultiLevel-NF~\cite{maeda2024frequency} & \textit{NF} & \textit{Single-cat.} & \xmark \\
    \textbf{Ours} & \textit{CL} & \textit{Multi-cat.} & \cmark \\
    \bottomrule
    \end{tabular}
\label{tab:comp-framework}
\end{table}

\subsection{Unified models for anomaly detection}
Unified models aim to detect anomalies across multiple categories using a single model, eliminating the need for category-specific retraining. Traditional AD methods~\cite{bergmann2020uninformed, defard2021padim, maeda2024frequency, liznerski2020explainable, you2022adtr, zavrtanik2021draem} typically adopt a one-model-per-category paradigm, which is effective when abundant training data is available but scales poorly in scenarios with many categories or scarce data.
To address these issues, You \textit{et al}.~\cite{you2022unified} proposed UniAD, a Transformer-based reconstruction framework equipped with three mechanisms: Layerwise Query Decoder, Neighbor Masked Attention, and Feature Jittering to prevent shortcut learning and improve cross-category performance.
Huang \textit{et al}.~\cite{huang2022registration} introduced RegAD, which employs a convolutional neural network (CNN) and spatial transformer network to align test samples with a support set, enabling category-agnostic inference without retraining.
Subsequent works have extended these ideas: OmniAL~\cite{Zhao_2023_CVPR} synthesizes pseudo-anomalies and uses a CNN-based architecture to jointly detect and localize anomalies across categories; HVQ-Trans~\cite{lu2023hierarchicalvectorquantizedtransformer} introduces hierarchical vector quantization to constrain reconstructions with category-specific prototypes; LafitE~\cite{yin2023lafitelatentdiffusionmodel} integrates latent diffusion models to avoid shortcut learning and enable feature-space editing; and Dinomaly~\cite{guo2025dinomaly} employs a ViT-based unified AD with a noisy bottleneck for robustness.
While unified AD has been studied in the image domain, it still remains rarely explored in the HAAD. We thus extend the unified model concept to skeleton-based HAAD, leveraging contrastive learning and generative motion augmentation to achieve robust few-shot AD without retraining.

\section{Method}
As shown in Fig.~\ref{fig:overview}, our framework consists of two modules, an action encoder and a generative model. The action encoder is optimized via contrastive learning to construct discriminative shared feature space. This capability enables few-shot HAAD by computing distance between support set and test samples in learned feature space. Here, the generative model serves as data augmentation to diversify training pair and support set.

Formally, let $\mathcal{X}^c = \{\mathbf{X}^c_1, \cdots, \mathbf{X}^c_{N_c}\}$ denote the set of human action sequences following the $c$-th action category, where $N_c$ denotes the number of samples in category $c$. Each action sequence is given by $\mathbf{X}_i = \{\mathbf{x}_1, \cdots, \mathbf{x}_H\} \in \mathbb{R}^{H \times 3J}$, where $\mathbf{x}_i$ refers to a pose in the $i$-th frame with $J$ three-dimensional joints. The goal of few-shot HAAD is to classify a test motion $\mathbf{X}^t$ as either normal or anomalous, with the use of an action set provided at inference time, which is termed as a support set $\mathcal{S}$. Here, the action category represented within the support set is defined as normal, while all other categories are regarded as anomalous. Specifically, the support set can be written as $\mathcal{S} = \{\mathbf{X}_1^s, \cdots, \mathbf{X}_{N_s}^s\}$, where each $\mathbf{X}_i^s$ serves as a reference normal action sample. As such, for a given action $\mathbf{X}^t$, we now aim to determine whether it belongs to the same category $s$ to inspect anomalies.

\begin{algorithm}[tb]
\caption{Optimization: Contrastive Learning}
\label{alg:1}
\begin{algorithmic}[1]
    \REQUIRE Training categories $\{1,\dots,N\}$; epochs $E$; number of generations $N_g$; pretrained \emph{frozen} HumanMAC $g(\cdot)$; action encoder $f_\theta(\cdot)$;  category function $c(\cdot)$
    \ENSURE Trained encoder $f_\theta$
    \FOR{epoch $=1,\ldots,E$}
        \STATE Sample $\mathcal{B} \gets \{\mathbf{X}_1,\dots,\mathbf{X}_{2N}\}$;
        \STATE Initialize $\mathcal{Z} \gets \emptyset$;
        \FOR{each $\mathbf{X}_i \in \mathcal{B}$}
            \STATE Generate $\{\tilde{\mathbf{X}}_{i,1},\dots,\tilde{\mathbf{X}}_{i,N_g}\} \gets g(\mathbf{X}_i)$;
            \STATE Encode $\boldsymbol{z}_{i,0} \gets f_\theta(\mathbf{X}_i)$;
            \STATE Encode $\boldsymbol{z}_{i,j} \gets f_\theta(\tilde{\mathbf{X}}_{i,j}), 
            \,\, j=1,\dots,N_g$;
            \STATE Append $\{\boldsymbol{z}_{i,0},\boldsymbol{z}_{i,1},
            \dots,\boldsymbol{z}_{i,N_g}\}$ to $\mathcal{Z}$;
        \ENDFOR
        \STATE Set $L \gets |\mathcal{Z}|$;
        \FOR{each $i \in \{1,\dots,L\}$}
            \STATE \COMMENT{positive pairs (same category, exclude self)}
            \STATE Find $\mathcal{P}(i) \gets \{ j \in \{1,\dots,L\} \setminus \{i\} \mid c(j) = c(i) \}$;
            \STATE Compute $\ell_i$ via Eq.~(\ref{eq:5});
        \ENDFOR
        \STATE Compute $\mathcal{L} \gets 1/L \sum_{i=1}^{L} \ell_i$;
        \STATE Update $\theta$ to minimize $\mathcal{L}$;
    \ENDFOR
\end{algorithmic}
\end{algorithm}

\begin{algorithm}[tb]
\caption{Inference: Few-shot HAAD}
\label{alg:2}
\begin{algorithmic}[1]
    \REQUIRE Test sample $\mathbf{X}^t$; support set $\mathcal{S}=\{\mathbf{X}_1,\dots,\mathbf{X}_{N_s}\}$; number of generations $N_g$; pretrained HumanMAC $g(\cdot)$; trained encoder $f_\theta(\cdot)$
    \ENSURE Anomaly score $a$   
    \STATE Initialize $\mathcal{V} \gets \emptyset$;
    \FOR{each $\mathbf{X}_i \in \mathcal{S}$}
        \STATE Generate $\{\tilde{\mathbf{X}}_{i,1},\dots,\tilde{\mathbf{X}}_{i,N_g}\} \gets g(\mathbf{X}_i)$;
        \STATE Encode $\boldsymbol{v}_{i,0} \gets f_\theta(\mathbf{X}_i)$;
        \STATE Encode $\boldsymbol{v}_{i,j} \gets f_\theta(\tilde{\mathbf{X}}_{i,j}),
        \,\, j=1,\dots,N_g$;
        \STATE Append $\{\boldsymbol{v}_{i,0}, \boldsymbol{v}_{i,1},\dots,\boldsymbol{v}_{i,N_g}\}$ to $\mathcal{V}$;
    \ENDFOR
    \STATE Encode $\boldsymbol{z} \gets f_\theta(\mathbf{X}^t)$;
    \STATE Compute $a$ via Eq.~(\ref{eq:6});
    \RETURN $a$
\end{algorithmic}
\end{algorithm}

\subsection{Action encoding}\label{Encoder}
Our method starts with the learning of action embeddings in the feature space that ideally support AD. To achieve this, we draw inspiration from \cite{mao2019learning} to perform first frequency domain encoding, and then design a Res-GCN to effectively encode the graph-structured human kinematics. 

\noindent \textbf{Frequency domain encoding.}  
In general, due to errors arising from sensors or 3D pose reconstruction, the recorded motion sequences often appear noisy and jittery. This can significantly damage the learning of true motion patterns to degrade the detection performance. To bypass this barrier, similar to \cite{NIPS2008_dc82d632}, we enforce the discrete cosine transform (DCT) to perform frequency domain encoding to better characterize action features. Specifically, the DCT for an arbitrary motion $\mathbf{X}$ is given by $\mathbf{C} = \mathbf{T} \mathbf{X}$, where $\mathbf{T} \in \mathbb{R}^{M \times H}$ is the predefined DCT basis matrix, and $\mathbf{C} \in \mathbb{R}^{M \times 3J}$ is the resulting DCT coefficient matrix. $M$ is the number of frequency components. Importantly, by retaining only low-frequency components, the DCT encoding contributes to effective noise suppression, yet also accurately reflects the original motion characteristics. We next need to learn the action dynamics within the coefficients $\mathbf{C}$.

\noindent \textbf{Graph convolutional network (GCN).}
Given that the human skeleton can be represented as a graph, we utilize the graph convolutional network (GCN) to learn spatial-temporal correlations among dynamic human joints. In particular, we define the GCN to contain $L$ layers, with $l \in \{1, \cdots, L\}$ denoting the layer index. Specifically, each GC layer updates node features $\mathbf{H}^{(l)} \in \mathbb{R}^{J \times F}$ at the $l$-th layer to: 
\begin{equation}
    \mathbf{H}^{(l+1)} = \sigma(\mathbf{A}^{(l)} \mathbf{H}^{(l)} \mathbf{W}^{(l)}),
\end{equation}
where  $\mathbf{H} ^{(l+1)} \in \mathbb{R}^{J \times \hat{F}}$ is the output feature, $\sigma(\cdot)$ is a non-linear activation, $\mathbf{A}^{(l)} \in \mathbb{R}^{J \times J}$ is a weighted adjacency matrix, and $\mathbf{W}^{(l)} \in \mathbb{R}^{F \times \hat{F}}$ is a trainable weight matrix. $F$ and $\hat{F}$ refer to the feature dimensionality before and after GC convolution at the $l$-th layer, respectively. Since our adjacency matrix is designed to be trainable rather than predefined as in \cite{kipf2016semi}, the GCN can adaptively explore and characterize complex dependencies across all joints during learning. We also design our GCN-based action encoder in a residual manner (Res-GCN) to ensure training stability under deeper architectures. Our GCN receives the DCT-transformed coefficients $\mathbf{C}$ as the input to the first layer to eventually produce the output feature $\mathbf{H} ^{(L+1)}$ that encodes action dynamics. 

As we aim to detect action anomalies, we further apply  max pooling over the node dimension of the final feature matrix $\mathbf{H} ^{(L+1)}$ to derive a vectorized representation 
$\boldsymbol{z} \in \mathbb{R}^{J}$, which compactly stores key action features to facilitate anomaly scoring.

\subsection{Generative motion augmentation}\label{HumanMAC}
The learning of our encoder is designed to follow a contrastive learning paradigm to facilitate learning discriminative representations. Typically, contrastive learning involves data augmentation to yield the robustness and representativeness of the encoded representations. To achieve this, we leverage HumanMAC \cite{Chen_2023_ICCV}, a diffusion-based generative model that formulates motion prediction as a masked motion completion task in the frequency domain. Specifically, HumanMAC predicts future motion frames conditioned on past observations. Given a motion sequence of length $H = O + P$, the model observes the first $O$ frames and generates the subsequent $P$ frames as the target segment. In our framework, HumanMAC serves two purposes: (1) to diversify training samples for better encoder generalization, and (2) to expand the support set during inference, improving robustness in few-shot matching. 

\noindent \textbf{Training of HumanMAC.}
HumanMAC is trained following the denoising diffusion probabilistic model ~\cite{ho2020denoising}, adapted to the frequency domain. Given a motion sequence $\mathbf{X}$, the DCT is applied to obtain the low-frequency spectrum $\mathbf{C}_0$. Then, Gaussian noise $\boldsymbol{\epsilon} \sim \mathcal{N}(0, I)$ is added at each timestep $t$ as:
\begin{equation}
    \mathbf{C}_t = \sqrt{\bar{\alpha}_t} \mathbf{C}_0 + \sqrt{1 - \bar{\alpha}_t} \boldsymbol{\epsilon},
\end{equation}
where $\bar{\alpha}_t = \prod_{i=1}^{t} \alpha_i$ denotes the cumulative noise schedule. A noise prediction network $\epsilon_\theta(\cdot, \cdot)$, consisting of a Transformer encoder and feed-forward layers, is trained to minimize the prediction error:
\begin{equation}
    \mathcal{L}_\text{noise} = \mathbb{E}_{\boldsymbol{\epsilon}, t} 
    \left[ \left\lVert \boldsymbol{\epsilon} - \epsilon_\theta(\mathbf{C}_t, t) \right\rVert^2 \right].
\end{equation}

\noindent \textbf{Conditional generation via DCT-completion.}
To enable conditional generation from observed sequence, HumanMAC adopts DCT-Completion, which integrates observation guidance into each denoising step. At every step $t$, two spectral estimates for timestep $t-1$ are computed: (1) a denoised spectrum $\mathbf{C}_{t-1}^d$ using $\epsilon_\theta(\mathbf{C}_t, t)$, and (2) a noisy spectrum $\mathbf{C}_{t-1}^n$ by adding Gaussian noise to the observed sequence. Both spectra are mapped to the temporal domain via iDCT and fused using a binary mask $\mathbf{M}$:
\begin{equation}
\resizebox{0.9\hsize}{!}{$
    \mathbf{C}_{t-1} = \mathbf{T} \left( \mathbf{M} \odot \mathbf{T}^{-1} \mathbf{C}_{t-1}^{n} + (1 - \mathbf{M}) \odot \mathbf{T}^{-1} \mathbf{C}_{t-1}^{d} \right),
    \label{eq:4}
    $}
\end{equation}
where $\mathbf{T}$ and $\mathbf{T}^{-1}$ denote the DCT and iDCT basis matrices, and $\odot$ indicates element-wise multiplication. The mask $\mathbf{M} \in \{0,1\}^{O+P}$ assigns 1 to observed frames (first $O$) and 0 to future frames (last $P$). This masked fusion preserves observed information and ensures controllable motion generation.

During both encoder training (Sec.~\ref{Optimization}) and inference (Sec.~\ref{Inference}), given an input sequence $\mathbf{X} = \{\mathbf{x}_1, \dots, \mathbf{x}_{O+P}\}$, HumanMAC observes the first $O$ frames to generate $N_g$ diverse completions for the remaining $P$ frames. This results in an augmented set $\tilde{\mathcal{X}} = \{\mathbf{\tilde{X}}_1, \dots, \mathbf{\tilde{X}}_{N_g}\}$ of realistic and semantically consistent motion sequences that also exhibit rich intra-category diversity.

\subsection{Optimization}\label{Optimization}
To enable our encoder (\textit{i.e.}, Res-GCN) to produce quality features to support AD, motivated by SimCLR~\cite{chen2020simple}, we leverage a contrastive learning framework to learn discriminative representations of human actions.
The basic contrastive learning framework encourages the encoder to bring positive pairs closer and push negative pairs apart in the representation space. Let the training set contain $N$ action categories. We construct a minibatch $\mathcal{B} = \{\mathbf{X}_1, \cdots, \mathbf{X}_{2N}\}$ at each training iteration by randomly sampling two instances from each category. Within each minibatch, a pair of samples from the same category is treated as a positive pair, while all others are treated as negative pairs. 

As discussed in Sec.~\ref{HumanMAC}, we introduce generative data augmentation to diversify the motion samples such that the generalizability of our model can be well enriched. Specifically, the generative model produces $N_g$ synthetic motion sequences for each sample in the minibatch. We can thus obtain an augmented minibatch  $\tilde{\mathcal{B}} = \{\{\mathbf{X}_1, \tilde{\mathcal{X}}_1\}, \cdots, \{\mathbf{X}_{2N}, \tilde{\mathcal{X}}_{2N}\}\}$, where each $\tilde{\mathcal{X}}_i$ contains $N_g$ samples generated from $\mathbf{X}_i$. Let $(i,j)$ denote an arbitrary index pair corresponding to a positive pair in the original minibatch $\mathcal{B}$. From the augmented minibatch $\tilde{\mathcal{B}}$, we define the set of extended positive pairs as all pairwise combinations between elements of $\{\mathbf{X}_i\} \cup \tilde{\mathcal{X}}_i$ and $\{\mathbf{X}_j\} \cup \tilde{\mathcal{X}}_j$. Then, the encoder takes the minibatch $\tilde{\mathcal{B}}$ and outputs representation vectors $\mathcal{Z} = \{\boldsymbol{z}_1, \cdots, \boldsymbol{z}_L\}$, where $L = 2N \cdot (1 + N_g)$ denotes the number of samples in $\tilde{\mathcal{B}}$. To incorporate the generated samples into contrastive learning, we extend the SimCLR loss formulation to handle multiple positive counterparts. Let $\mathcal{P}(i)$ denote the set of indices corresponding to positive samples for a given sample $i$. The contrastive loss for sample $i$ can be given by:
\begin{equation}
\resizebox{0.9\hsize}{!}{$
  \ell(i) = \sum_{j \in \mathcal{P}(i)} -\log \frac{\exp\left( \mathrm{sim}(\boldsymbol{z}_i, \boldsymbol{z}_j)/\tau \right)}
  {\sum_{k=1}^{L} \mathbbm{1}_{[k \ne i]} \exp\left( \mathrm{sim}(\boldsymbol{z}_i, \boldsymbol{z}_k)/\tau \right)},
  $}
  \label{eq:5}
\end{equation}
where $\mathrm{sim}(\cdot)$ denotes the cosine similarity and $\tau$ is a temperature scaling parameter. The numerator promotes similarity with positive counterparts, while the denominator includes all other samples as negatives (excluding $i$ itself). The final loss is obtained by averaging over all samples in the minibatch:
$\mathcal{L} =1/L \sum_{i=1}^{L} \ell(i)$.
For clarity, the overall optimization flow is summarized in Algorithm~\ref{alg:1}.

\subsection{Inference}\label{Inference}
During inference, we utilize the support set to realize few-shot HAAD. Importantly, the support set $\mathcal{S}$ serves as a baseline of normality against which test samples are compared. To enrich the diversity of this limited baseline, we similarly employ the generative motion augmentation to produce $N_g$ synthetic samples for each $\mathbf{X}_i^s$, resulting in an augmented support set $\tilde{\mathcal{S}} = \{ \mathbf{X}^s_1, \cdots, \mathbf{X}^s_A\}$ with  $A= N_s\cdot(1 + N_g)$ samples in total. This augmentation strategy aims to mitigate the sensitivity of AD performance to the particular composition of the support set by introducing semantically consistent intra-category variations. For simplicity, we define $\mathbf{X}^t$ as a single test sample. Both the augmented support set and the test sample are passed through the encoder to obtain their respective feature representations: $\mathcal{V} = \{\boldsymbol{v}_1, \cdots, \boldsymbol{v}_{A}\}$ for the support set $\tilde{\mathcal{S}}$, and $\boldsymbol{z}$ for the test sample $\mathbf{X}^t$. To evaluate how anomalous the test sample is, we compute the average Euclidean distance between the test feature vector $\boldsymbol{z}$ and each support vector $\boldsymbol{v}_i \in \mathcal{V}$. Specifically, the anomaly score $a$ is defined as:
\begin{equation}
a = \frac{1}{A} \sum_{i=1}^{A} \lVert \boldsymbol{z} - \boldsymbol{v}_i \rVert_2,
\label{eq:6}
\end{equation}
where a larger value of $a$ indicates a higher degree of anomaly. Because the encoder is trained to produce compact and discriminative representations via contrastive learning, the distance in this space effectively captures the semantic similarity between actions. This discriminative property enables the model to detect anomalies even in previously unseen action categories, as long as the support set contains samples from the corresponding category and serves as the normal reference. 
For clarity, the inference flow is summarized in Algorithm~\ref{alg:2}.

\section{Experiment}
\subsection{Experimental setup}
\noindent\textbf{Dataset.}
We evaluate the proposed method on HumanAct12~\cite{guo2020action2motion}, which consists of 12 daily human activity categories. Each motion sample provides 3D coordinates of 24 skeletal joints. Totally, HumanAct12 contains 1191 motion clips with 90,099 frames.

\noindent\textbf{Metrics.}
To evaluate the anomaly detection accuracy, we employ the Area Under the Receiver Operating Characteristic Curve (AUC), which serve as a threshold-agnostic performance metrics.

\noindent\textbf{Implementation details.}
The action encoder is optimized using the Adam optimizer~\cite{kingma2014adam} for 100 epochs, with a learning rate linearly decaying from 1e-3 to 1e-5. The temperature parameter $\tau$ for contrastive loss in Eq.~(\ref{eq:5}) is set to 1. We use ten DCT bases and adopt a Res-GCN~\cite{mao2019learning} composed of four residual blocks, each with two GCN layers and a hidden dimension of 128.
To assess the model’s generalization ability to unseen categories, three action categories (\textit{Phone}, \textit{Boxing}, and \textit{Throw}) are excluded during training and used only for the testing phase. During testing, a small support set is provided for these categories. This setting simulates few-shot scenario, enabling evaluation of the model’s ability to detect anomalies in previously unseen actions.
The motion generation module (i.e., HumanMAC~\cite{Chen_2023_ICCV}) is pretrained on HumanAct12 for 1000 epochs using the DDIM sampling strategy~\cite{song2020denoising} with 100 diffusion steps. A cosine noise scheduler~\cite{nichol2021improved} is used to control the variance schedule. The number of DCT bases is set to 20 for HumanMAC, as in \cite{Chen_2023_ICCV}. The noise prediction network consists of eight Transformer-based blocks.
All motion sequences are pre-processed to have a fixed length of 60 frames. During both training and inference, we apply the same conditioning setup: the first 30 frames serve as the observed input and the remaining 30 frames as the target for prediction. 

\begin{table*}[tb]
\centering
\caption{\textbf{Quantitative results} of category-wise AUC scores on HumanAct12. $^{\dagger}$ indicates categories excluded from training and evaluated in a few-shot setting only for our method (Ours). Other methods were trained per category. We report the average and standard deviation of AUC on ten trials with different seeds. The best and second-best results are highlighted in bold and underlined, respectively.
}
\label{tab:comparison}
\begin{tabular}{lcccc}
    \toprule
    Category & STG-NF~\cite{hirschorn2023normalizing} & MoCoDAD~\cite{flaborea2023multimodal} & MultiLevel-NF~\cite{maeda2024frequency} & Ours \\
    \midrule
    \textit{Warm up}       & $0.720$ & $0.664$ & $\underline{0.842}$ & $\mathbf{0.891}\pm0.001$\\
    \textit{Walk}          & $\underline{0.835}$ & $0.627$ & $\mathbf{0.907}$ & $0.758\pm0.003$\\
    \textit{Run}           & $0.568$ & $0.516$ & $\mathbf{0.781}$ & $\underline{0.614}\pm0.015$\\
    \textit{Jump}          & $0.861$ & $0.475$ & $\underline{0.912}$ & $\mathbf{0.973}\pm0.002$\\
    \textit{Drink}         & $0.751$ & $0.631$ & $\underline{0.776}$ & $\mathbf{0.871}\pm0.004$\\
    \textit{Lift dmbl}     & $0.926$ & $0.479$ & $\mathbf{0.958}$ & $\underline{0.947}\pm0.000$\\
    \textit{Sit}           & $0.878$ & $0.406$ & $\mathbf{0.904}$ & $\underline{0.899}\pm0.000$\\
    \textit{Eat}           & $\underline{0.972}$ & $0.393$ & $\mathbf{0.986}$ & $0.965\pm0.001$\\
    \textit{Trn steer whl} & $0.855$ & $0.480$ & $\underline{0.869}$ & $\mathbf{0.883}\pm0.007$\\
    \textit{Phone}$^{\dagger}$  & $\underline{0.791}$ & $0.406$ & $0.783$ & $\mathbf{0.921}\pm0.004$\\
    \textit{Boxing}$^{\dagger}$ & $0.602$ & $0.358$ & $\underline{0.619}$ & $\mathbf{0.883}\pm0.015$\\
    \textit{Throw}$^{\dagger}$  & $0.766$ & $0.497$ & $\underline{0.782}$ & $\mathbf{0.824}\pm0.026$\\
    \midrule
    Mean          & $0.794$ & $0.494$ & $\underline{0.843}$ & $\mathbf{0.869}\pm0.006$\\
    \bottomrule
\end{tabular}
\vspace{0.5em}
\end{table*}

\begin{figure*}[tb]
    \centering
    \subfloat[\textit{Jump}\label{fig:jump}]{
        \includegraphics[clip, width=1.0\columnwidth]{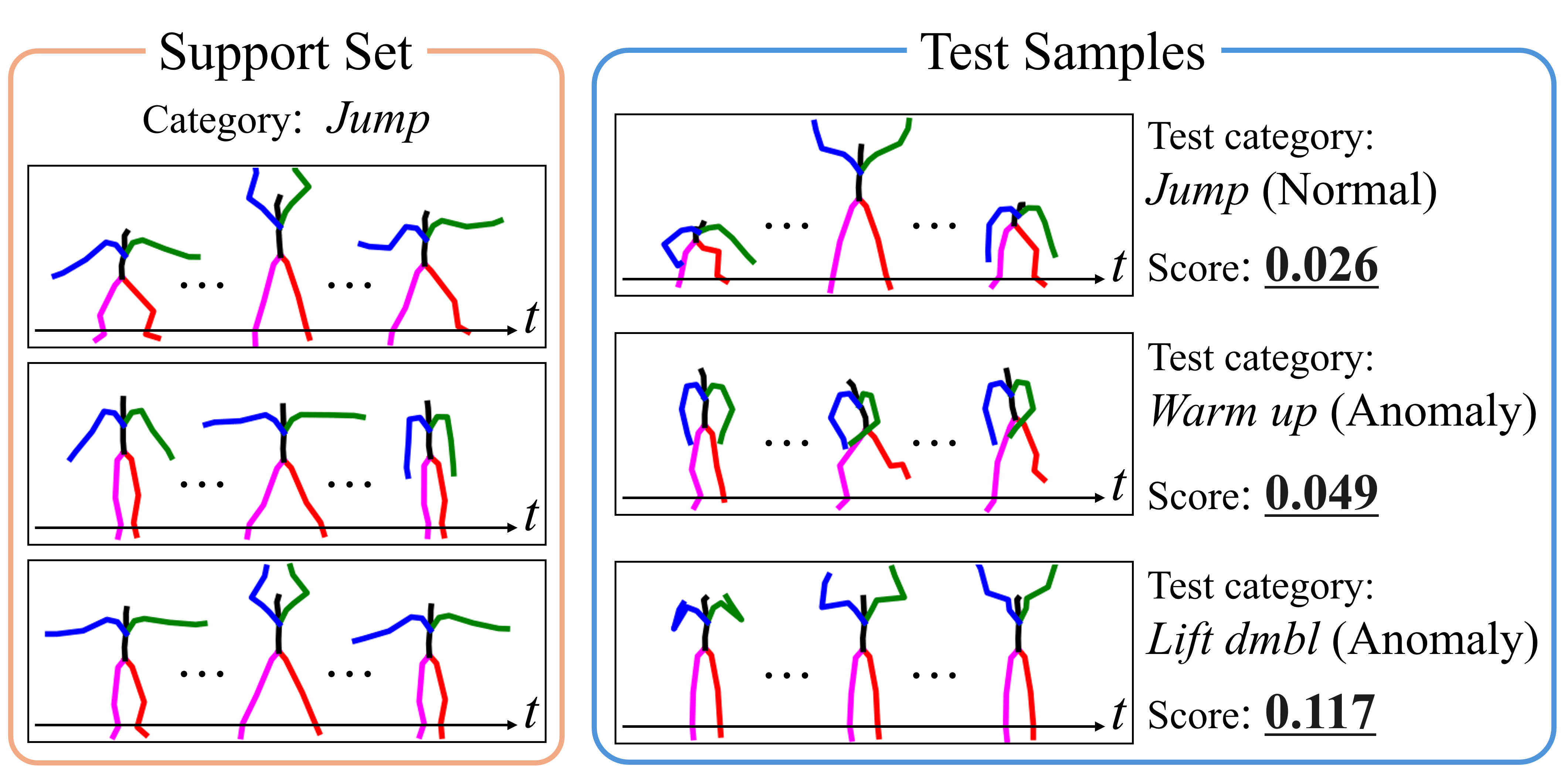}
    }
    \hfill
    \subfloat[\textit{Phone}\label{fig:phone}]{
        \includegraphics[clip, width=1.0\columnwidth]{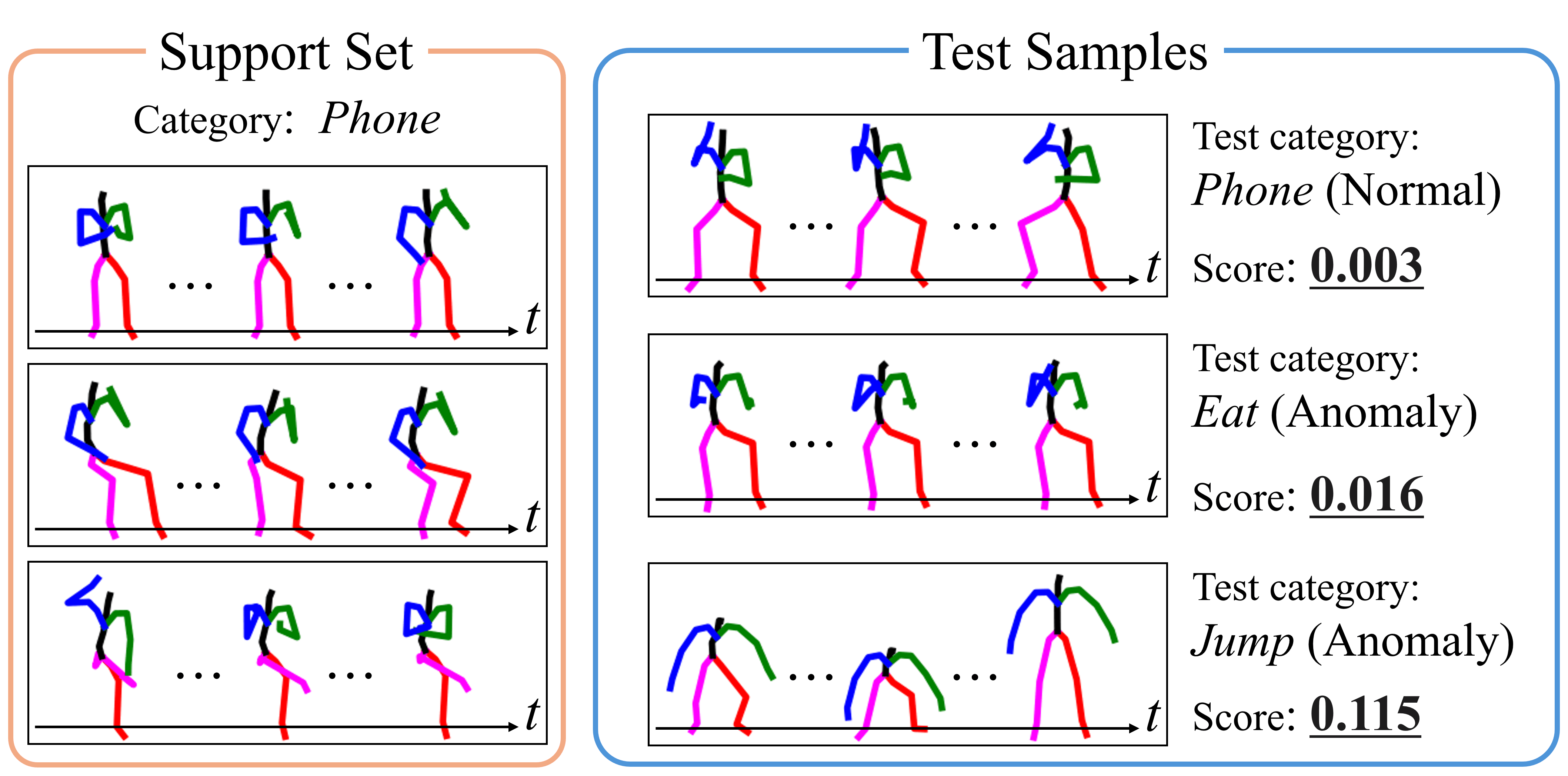}
    }
    \caption{\textbf{Qualitative results}, with the normal category set to \textit{Jump} (whole-body action) in (a),  and \textit{Phone} (part-body action) in (b) for few-shot HAAD. Our few-shot AD framework allows for the use of a support set (orange box) to score given test samples (blue box). Higher scores indicate a greater degree of anomalies.
    }
    \label{fig:action}
\end{figure*}

\subsection{Comparison with state-of-the-art methods}
\noindent\textbf{Quantitative evaluation.}
To verify the effectiveness of our unified framework, we compare the proposed unified model with three state-of-the-art HAAD methods. The baselines include STG-NF~\cite{hirschorn2023normalizing}, MoCoDAD~\cite{flaborea2023multimodal}, and MultiLevel-NF~\cite{maeda2024frequency}, all of which adopt a \emph{one-model-per-category} strategy, where a separate model is trained for each action category.
As shown in Tab.~\ref{tab:comparison}, compared with the three baselines, the proposed method achieves the highest mean AUC of $0.869$ under the more challenging few-shot setting. 
In particular, our method attains the best performance in 7 out of 12 categories, including not only seen categories but also unseen categories (\textit{Phone}, \textit{Boxing}, and \textit{Throw}).
These results demonstrate robustness across diverse motion types and confirm strong generalization ability.
Overall, by training once on multiple categories and detecting anomalies in both seen and unseen actions under a few-shot setting, our unified framework offers a practical and retraining-free solution for HAAD.

\begin{figure*}[tb]
    \centering
    \subfloat[Representations learned without generative augmentation\label{fig:tsne-without}]{
        \includegraphics[clip, width=0.98\columnwidth]{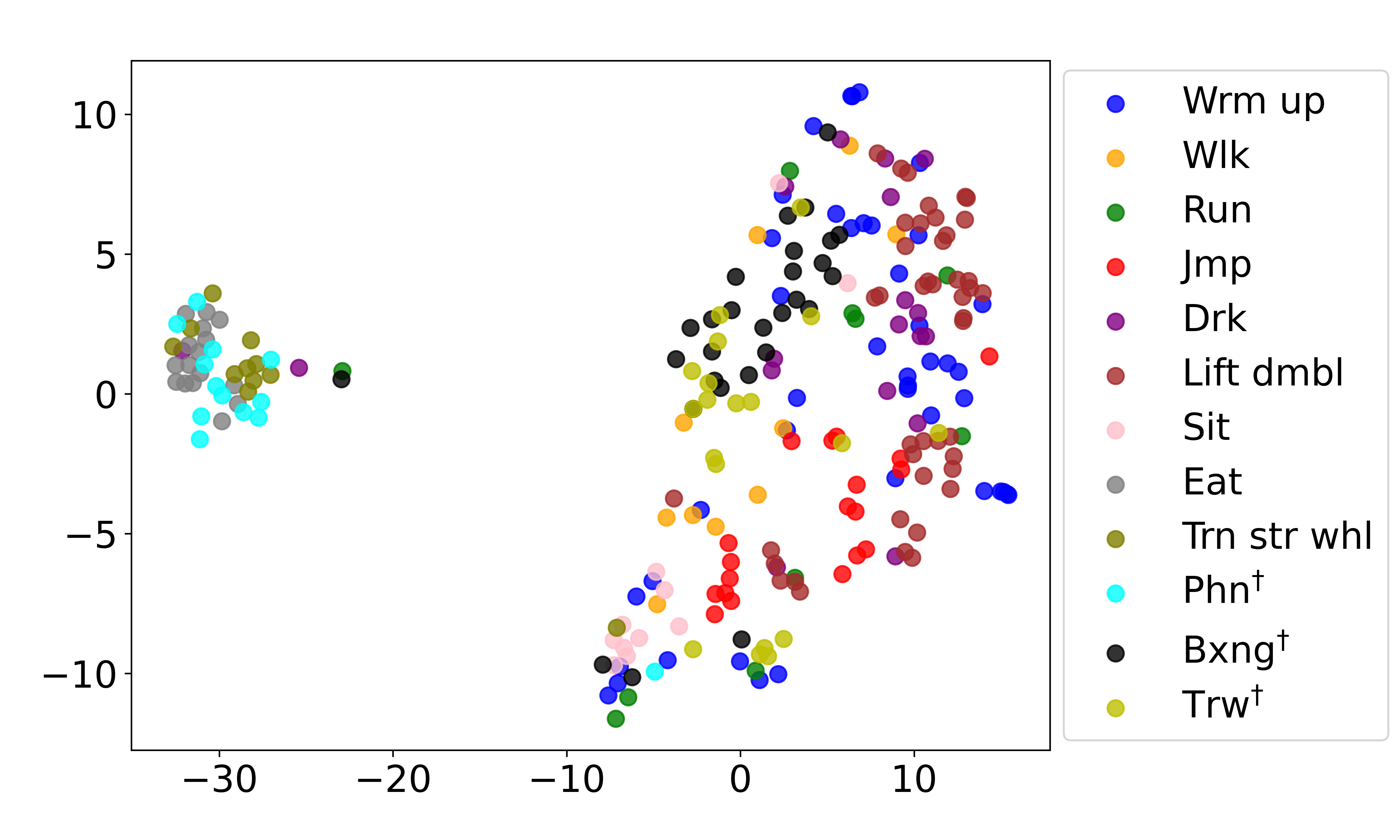}
    }
    \hfill
    \subfloat[Representations learned with generative augmentation\label{fig:tsne-with}]{
        \includegraphics[clip, width=0.98\columnwidth]{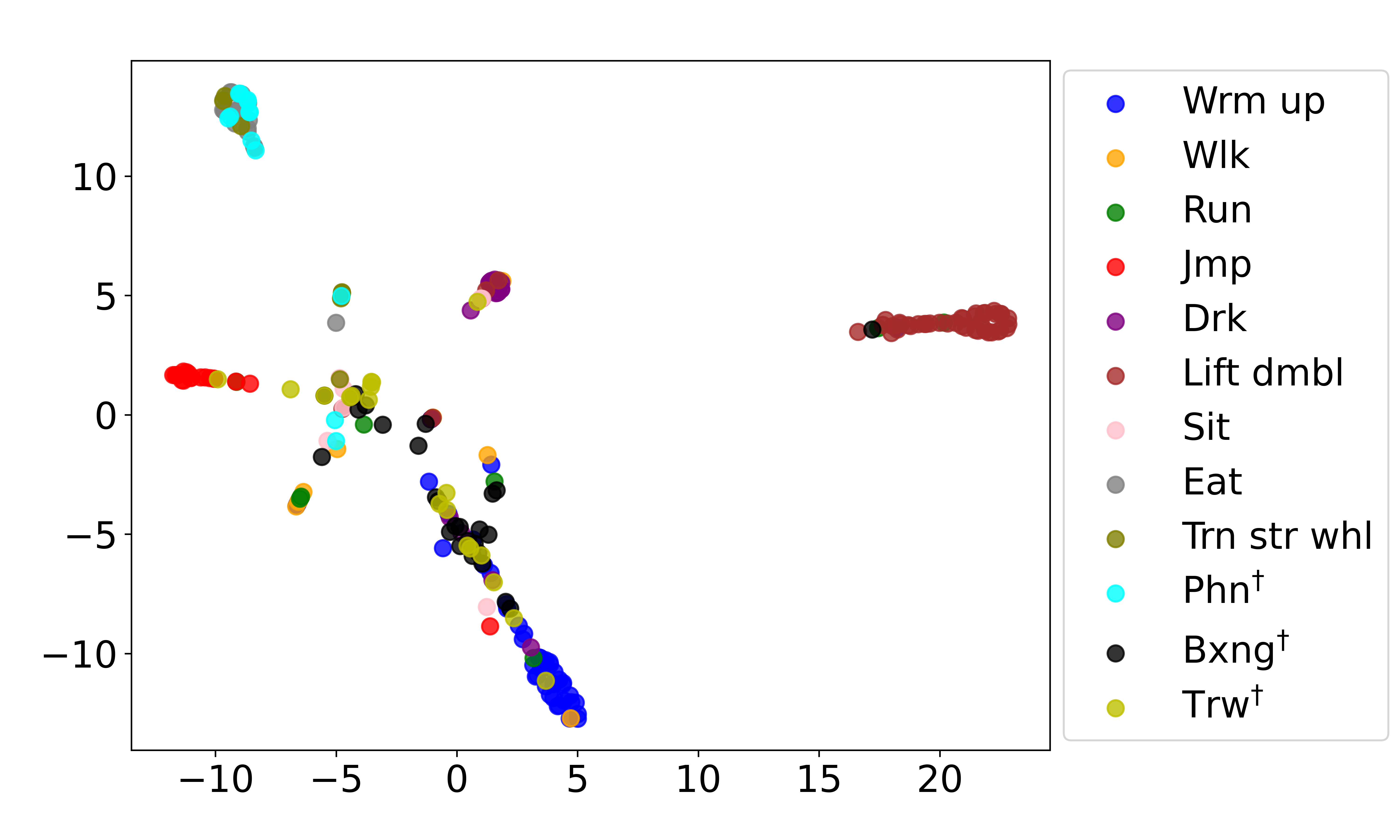}
    }
    \caption{\textbf{t-SNE visualization} of 
    feature space learned (a) without and (b) with generative augmentation during contrastive learning.  }
    \label{fig:tsne}
\end{figure*}

\noindent\textbf{Qualitative evaluation.}
To give an intuitive understanding of how anomaly scores are distributed across different test actions, we visualize representative samples under a fixed support set in Fig.~\ref{fig:action}. In particular, Fig.~\ref{fig:jump} and Fig.~\ref{fig:phone} illustrate the cases where \textit{Jump} and \textit{Phone} are set as the normal categories, respectively.
Overall, in both cases, the normal category consistently receives the lowest anomaly score, while anomalous categories yield higher scores depending on their similarity to the support samples.
Notably, when \textit{Jump} is set as the normal category, the score of \textit{Warm up} is lower than that of \textit{Lift dmbl}. It is worth noting that \textit{Warm up} is also a whole-body action similar to \textit{Jump}, while \textit{Lift dmbl} involves only part-body movement.
In contrast, when the support set is constructed from the part-body action \textit{Phone} (Fig.~\ref{fig:phone}), the part-body action \textit{Eat} receives a lower score, whereas the whole-body action \textit{Jump} yields a much higher score.
These observations indicate that the learned encoder constructs a discriminative representation space where semantically similar actions are embedded closer together, whereas dissimilar actions are placed farther apart. 
Such a discriminative space is well suited for distinguishing normal and anomalous actions in few-shot HAAD tasks.

\subsection{Ablation study}
\noindent\textbf{Quantitative evaluation.}
To validate the effectiveness of our generative motion augmentation strategy, we conduct an ablation study by selectively enabling or disabling generation during training-time and inference-time. Tab.~\ref{tab:ablation} summarizes the results for each configuration.
Compared to the baseline without augmentation (1st row), inference-time augmentation alone (2nd row) does not improve detection accuracy, but it reduces variance across runs. This suggests that augmenting support set contributes to stability against variations in the support set. 
In contrast, training-time augmentation alone (3rd row) yields a clear improvement in accuracy over the baseline (1st row). 
A possible explanation is that generative augmentation increases the diversity of contrastive pairs, which in turn contributes to better discriminative performance.
Consequently, using augmentation in both training and inference (4th row) achieves the best overall performance.
This finding suggests that the two augmentation strategies are complementary: training-time augmentation improves discriminative capacity, while inference-time augmentation improves stability.
Therefore, their combination provides the most practical and robust solution for few-shot HAAD.

\begin{table}[tb]
\centering
\caption{\textbf{Augmentation strategy ablation} on HumanAct12 with AUC.
“aug.” denotes generative motion augmentation.}
\label{tab:ablation}
\small
\begin{tabular}{ccc}
    \toprule
    Training-time aug. & Inference-time aug. & AUC\\
    \midrule
    \xmark & \xmark & $0.760\pm0.044$\\
    \xmark & \cmark & $0.754\pm0.038$\\
    \cmark & \xmark & $0.866\pm0.012$\\
    \cmark & \cmark & $\mathbf{0.869}\pm\mathbf{0.006}$\\
    \bottomrule
\end{tabular}
\end{table}

\noindent\textbf{Qualitative evaluation.}
To further investigate the impact of generative augmentation in contrastive learning, we visualize the representations of test samples using t-SNE~\cite{maaten2008visualizing}, as shown in Fig.~\ref{fig:tsne}.
It can be seen that the representations learned without augmentation (Fig.~\ref{fig:tsne-without}) form coarse two groups of clusters: actions involving part-body movements (e.g., \textit{Eat}, \textit{Trn steer whl}, \textit{Phone}) are roughly separated from whole-body motions (e.g., \textit{Warm up}, \textit{Walk}, \textit{Jump}), but categories remain entangled within each group.
In contrast, with augmentation (Fig.~\ref{fig:tsne-with}), the clusters show clearer category-level separation, and even within whole-body actions such as \textit{Warm up} and \textit{Jump}, categories become better separated.
These results suggest that generative augmentation increases the diversity of contrastive pairs, 
enabling the encoder to learn more discriminative action representations.
Consequently, the model effectively captures differences between each category, leading to improved HAAD accuracy.

\subsection{Parameter sensitivity experiments}
This section evaluates the sensitivity of model performance to key hyper-parameters, including the number of support samples at inference, the number of generated samples during contrastive learning, and the prediction length in the generative model.

\begin{figure}[tb]
    \centering
    \includegraphics[width=0.98\linewidth]{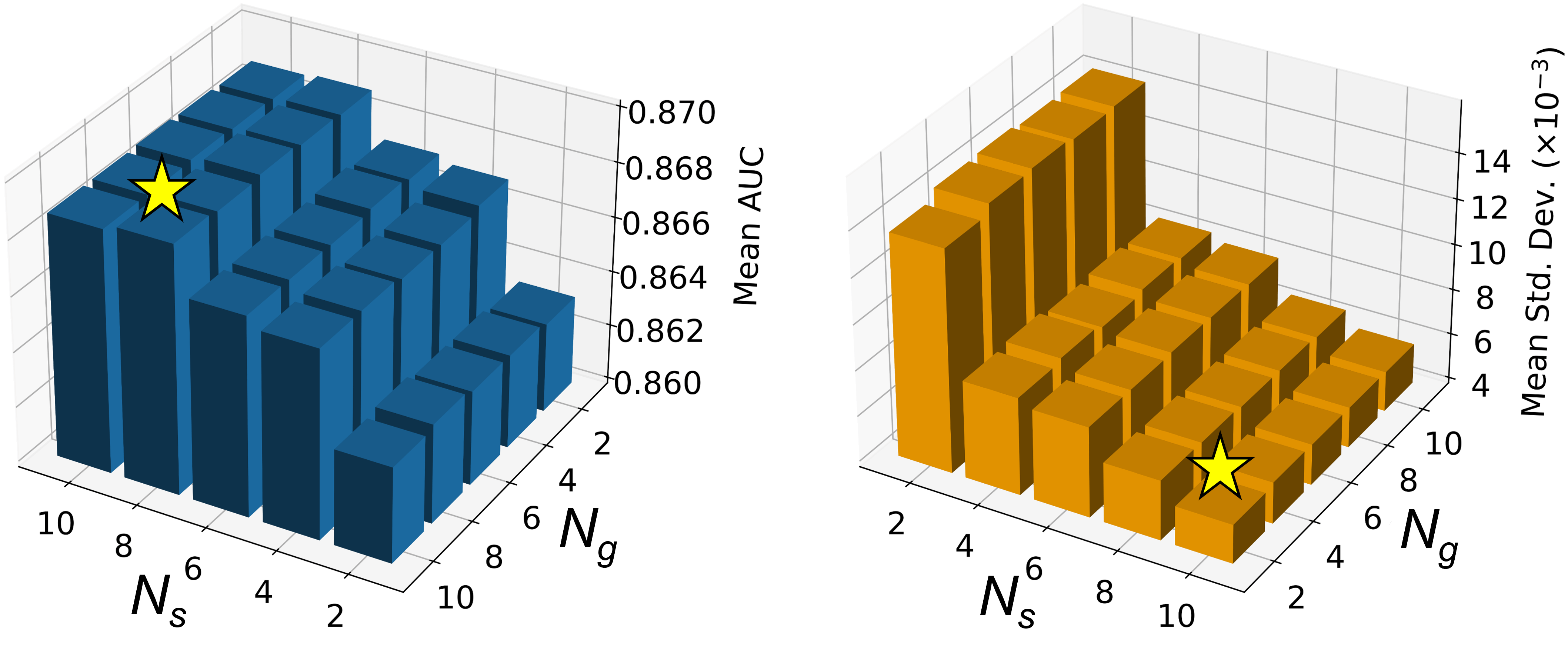}
    \caption{
    \textbf{Performance analysis}  on the number of support samples at inference. The left figure shows the mean AUC scores, and the right shows the mean standard deviation. $N_s$ denotes the number of support samples per category, and $N_g$ is the number of generated samples per support sample. Star ($\star$) localizes the best performance: the highest mean AUC score (left) and the lowest mean standard deviation (right).}
    \label{fig:support_num}
\end{figure}

\begin{figure}[tb]
    \centering
    \includegraphics[width=1.0\linewidth]{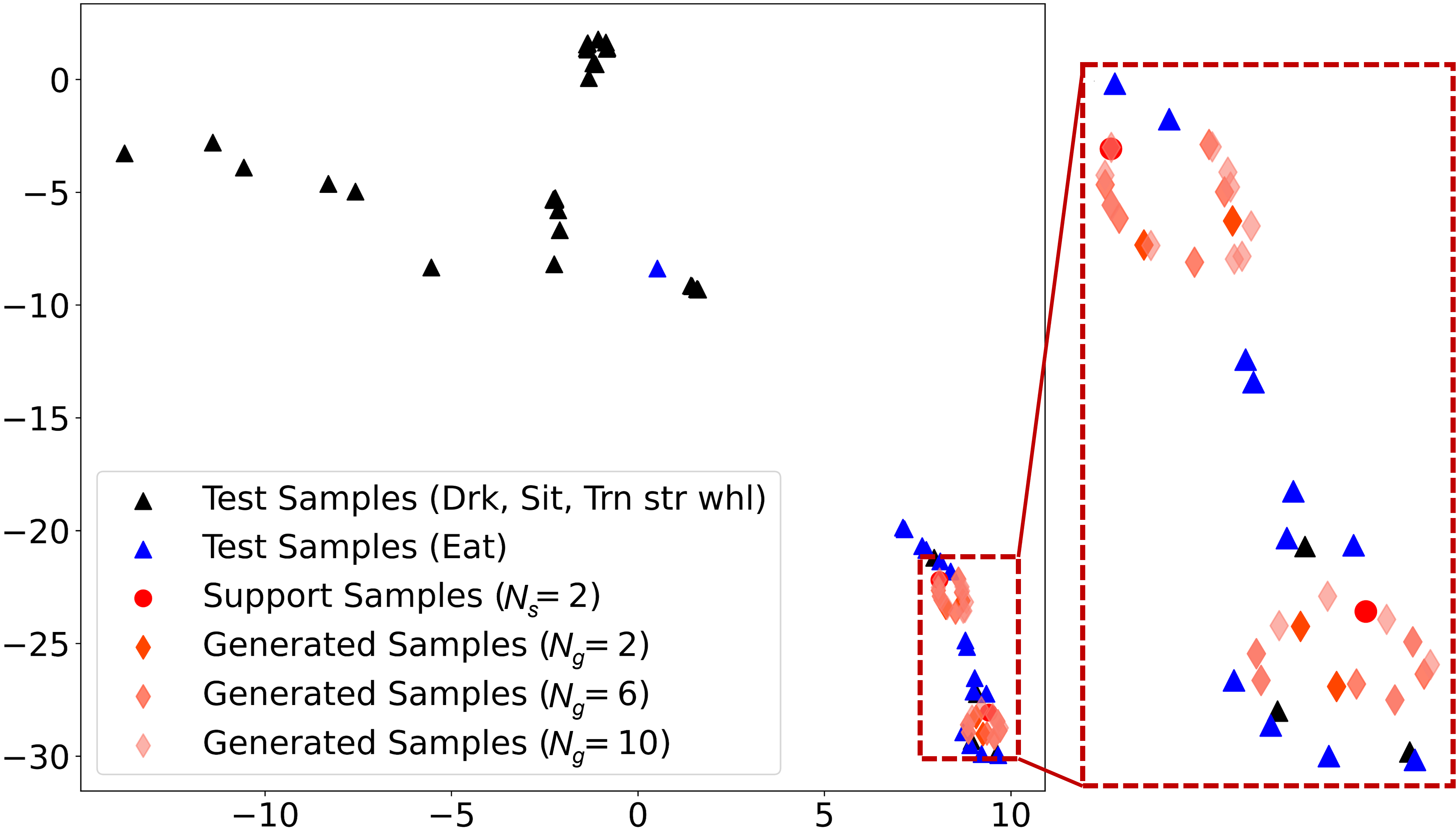}
    \caption{\textbf{t-SNE visualization} of the feature space on different categories and sample number settings. The normal category is set to \textit{Eat}. $N_s$ denotes the number of real support samples, and $N_g$ is the number of generated samples per support sample.}
    \label{fig:tsne-generated}
\end{figure}

\noindent\textbf{The number of support samples.}
The proposed method detects anomalies by calculating the distance between test samples and support samples in the learned representation space. 
Therefore, we first investigate how the number of real support samples ($N_s$) and generated samples per real sample ($N_g$) affect AD performance.
As shown in Fig.~\ref{fig:support_num}, increasing $N_s$ consistently improves the average AUC score and reduces the standard deviation. In contrast, increasing $N_g$ brings only marginal gains.
To better understand this behavior, Fig.~\ref{fig:tsne-generated} visualizes real support samples, generated samples, and test samples. The visualization reveals that the generated samples lie close to their corresponding real samples, forming tight clusters. 
This observation suggests that contrastive learning encourages synthetic samples to align with real ones, which may limit the additional diversity gained from support set augmentation.
In summary, the number of generations $N_g$ 
do not impose noticeable influence on the AD  performance, and instead, real samples play the dominant role in few-shot HAAD.

\begin{figure}
    \centering
    \includegraphics[width=1.0\linewidth]{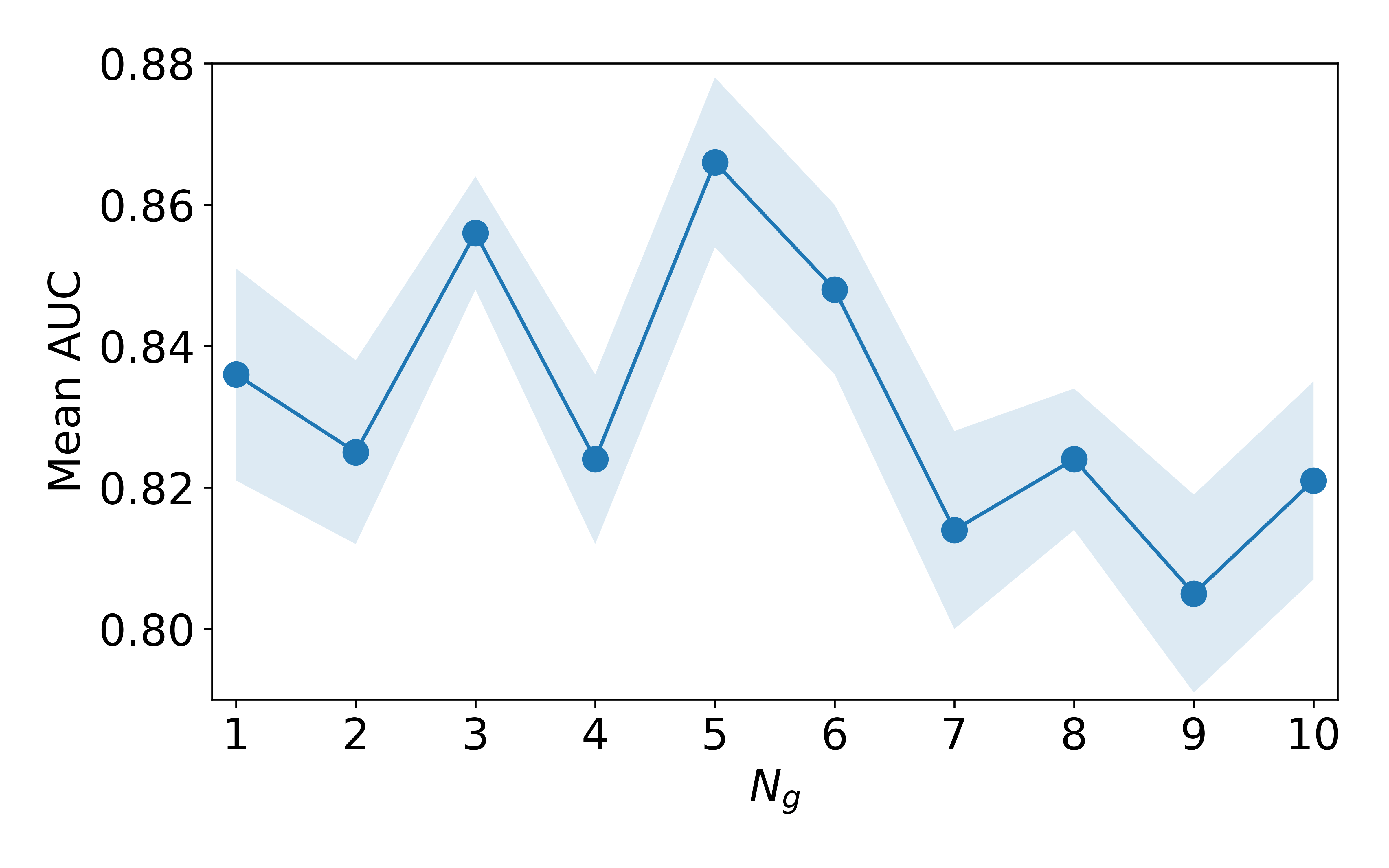}
    \caption{\textbf{Performance analysis} on the number of samples $N_g$ during training. For every sample number setting, we show average performance and the mean standard deviation on all categories.}
    \label{fig:cl_num}
\end{figure}

\noindent\textbf{The number of generations in training.}
We introduce generative augmentation during training to enrich contrastive pairs with diverse yet semantically consistent samples. 
To evaluate the effectiveness of this strategy, we next investigate how the number of generated samples per real sample ($N_g$) during training influences AD performance.
During evaluation, the support set contains three real samples per category without data augmentation at inference.
Fig.~\ref{fig:cl_num} summarizes the AUC scores when $N_g$ is varied from 1 to 10. 
The value of AUC improves when $N_g$ is increased up to around 3–5, but degrades when $N_g$ becomes larger. 
A possible reason is that excessive generations introduce redundancy to contrastive pairs and reduce discriminative effectiveness.
The above analysis suggests that the benefit of generative augmentation lies in balancing diversity and redundancy, rather than simply increasing the number of generations.

\begin{table}[tb]
\centering
\caption{\textbf{Quantitative results} of AUC scores on HumanAct12 under varying configurations of observed and predicted frame lengths for support set augmentation.}
\label{tab:obs/pre}
\begin{tabular}{ccc}
    \toprule
    Observed length & Predicted length & AUC \\
    \midrule
    $50$ & $10$ & $0.866 \pm \mathbf{0.012}$ \\
    $40$ & $20$ & $\mathbf{0.867} \pm \mathbf{0.012}$ \\
    $30$ & $30$ & $\mathbf{0.867} \pm \mathbf{0.012}$ \\
    $20$ & $40$ & $0.866 \pm 0.014$ \\
    $10$ & $50$ & $0.861 \pm 0.021$ \\
    \bottomrule
\end{tabular}
\end{table}

\noindent\textbf{The prediction length.}
The generative model in our framework predicts $P$ future motion frames from $O$ observed frames, thereby providing synthetic sequences for augmentation. 
To study the effect of the prediction length $P$, we vary the ratio of $O$ to $P$ while fixing the total length at 60. Here, the support set consists of three original samples per category, with 10 synthetic samples generated per original at inference.
Tab.~\ref{tab:obs/pre} shows that performance is highest when $P$ is set to 20–30, but degrades with longer horizons such as $P=50$.
This degradation may be attributed to two factors: (1) shorter $O$ provides limited contextual information to guide generation, and (2) longer $P$ introduces greater uncertainty.
Therefore, based on this observation, we set $P=30$ as the default.

\section{Conclusion}
In this paper, we propose a unified similarity-based framework for few-shot HAAD. Unlike existing methods that require category-specific retraining, our approach enables AD in both seen and unseen action categories by using only a small support set. By constructing a category-agnostic representation space through contrastive learning and leveraging generative motion augmentation via HumanMAC, our model effectively captures both inter-category discriminability and intra-category variability. Extensive experiments on the HumanAct12 dataset demonstrate that our framework achieves state-of-the-art performance across various motion types while significantly improving generalizability and scalability in data-scarce scenarios. These results validate the effectiveness of our approach as a practical solution for the few-shot HAAD scenario.

While our framework achieves satisfactory performance, it assumes fixed input structures such as the number of joints and skeletal topology. In practice, motion data can vary across sensors and datasets. In the future, we aim to extend our model to handle heterogeneous inputs with flexible graph structures to further boost robustness.

\section*{Acknowledgments}
This work is supported by JSPS KAKENHI Grant Number 23K10712.

{
    \small
    \bibliographystyle{unsrt}
    \bibliography{main}
}
\end{document}

%% file: preamble.tex
\usepackage{tabu}
\usepackage{booktabs}
\usepackage{times}
\usepackage{epsfig}
\usepackage{graphicx}
\usepackage{amsmath}
\usepackage{amssymb}
\usepackage{color}
\usepackage{algorithmic}
\usepackage{algorithm}

\usepackage{bbm}
\usepackage{bm}
\usepackage{multirow}
\usepackage{comment}
\usepackage{here}
\usepackage{arydshln}
\usepackage{multicol}
\usepackage{iccv}
\usepackage{textcomp}
\usepackage{tipa}
\usepackage{url}
\usepackage{float}
\usepackage{appendix} 
\usepackage{tabularx}
\usepackage{pgffor}
\usepackage{pifont}
\newcommand{\cmark}{\ding{51}} 
\newcommand{\xmark}{\ding{55}} 